\pdfoutput=1

\documentclass[11pt]{article}

\usepackage[preprint]{acl}

\usepackage{times}
\usepackage{latexsym}

\usepackage[T1]{fontenc}

\usepackage[utf8]{inputenc}

\usepackage{microtype}

\usepackage{inconsolata}

\usepackage{array}
\usepackage{pifont}
\usepackage{tabularx}
\usepackage{adjustbox}
\usepackage{multirow}
\usepackage{enumitem}
\usepackage{xspace}
\usepackage{tcolorbox}
\usepackage{booktabs,amsfonts,dcolumn}
\usepackage{hyperref}
\usepackage{url}
\usepackage{amsmath,amsthm,amsfonts,amssymb,bm,stmaryrd,bbm}
\usepackage[noorphans,vskip=0.75ex,leftmargin=2ex]{quoting}
\usepackage{footmisc}\interfootnotelinepenalty=10000
\usepackage{colortbl}
\usepackage{cleveref}
\usepackage{textcomp}
\usepackage{afterpage,caption}
\usepackage{comment}

\newcommand{\cmark}{\ding{51}}
\newcommand{\xmark}{\ding{55}}

\newcommand\tf[1]{\textbf{#1}}
\newcommand\ttt[1]{\texttt{#1}}

\renewcommand{\paragraph}[1]{\vspace{0.2cm}\noindent\textbf{#1}}

\newcommand{\tableindent}{~~}

\newcommand{\ours}{PTP}
\newcommand{\bi}{PTP} %

\newcommand{\improve}{{8}}

\newcommand{\maskrate}{10\%}

\newcommand{\bsl}{\makebox[0pt][r]{\raisebox{0.05em}{$\bigstar\,$}}}

\definecolor{cello}{HTML}{f2f3f5}
\newcommand{\colorcello}{\cellcolor{cello}}

\title{
    Improving Language Understanding from Screenshots %
}

\author{Tianyu Gao\quad Zirui Wang\quad Adithya Bhaskar\quad Danqi Chen\\
Princeton Language and Intelligence (PLI), 
Princeton University \\
\ttt{\{tianyug,zw1300,ab4197,danqic\}@cs.princeton.edu}\quad
}

\begin{document}
\maketitle

\begin{abstract}

An emerging family of language models (LMs),
capable of processing both text and images within a single visual view,
has the promise to unlock complex tasks such as chart understanding and UI navigation.
We refer to these models as \textit{screenshot language models}. 
Despite their appeal, %
existing screenshot LMs  
substantially lag behind text-only models on language understanding tasks.
To close this gap, 
we adopt a simplified setting where the model inputs are plain-text-rendered screenshots,
and we focus on improving the text ability of screenshot LMs.
We propose a novel 
\tf{P}atch-and-\tf{T}ext \tf{P}rediction
(\textbf{\ours{}}) objective,
which 
masks and recovers
both 
image patches of screenshots
and text within screenshots. 
We also conduct extensive ablation studies on masking rates and patch sizes, as well as designs for improving training stability.
Our pre-trained model, %
while solely taking visual inputs,
achieves comparable performance with BERT on 6 out of 8 GLUE tasks (within 2\%)  and 
improves up to \improve{}\% over prior work. 
Additionally, we extend \ours{} to train autoregressive screenshot LMs
and demonstrate 
its effectiveness---our models can significantly reduce perplexity by utilizing the screenshot context.
Together, we hope our findings can inspire future research on developing powerful screenshot LMs and extending their reach to broader applications.\footnote{Code: \url{https://github.com/princeton-nlp/PTP}.}

\end{abstract}

\section{Introduction}
\label{sec:intro}

\begin{figure}[t]
    \center
    \includegraphics[width=0.98\linewidth]{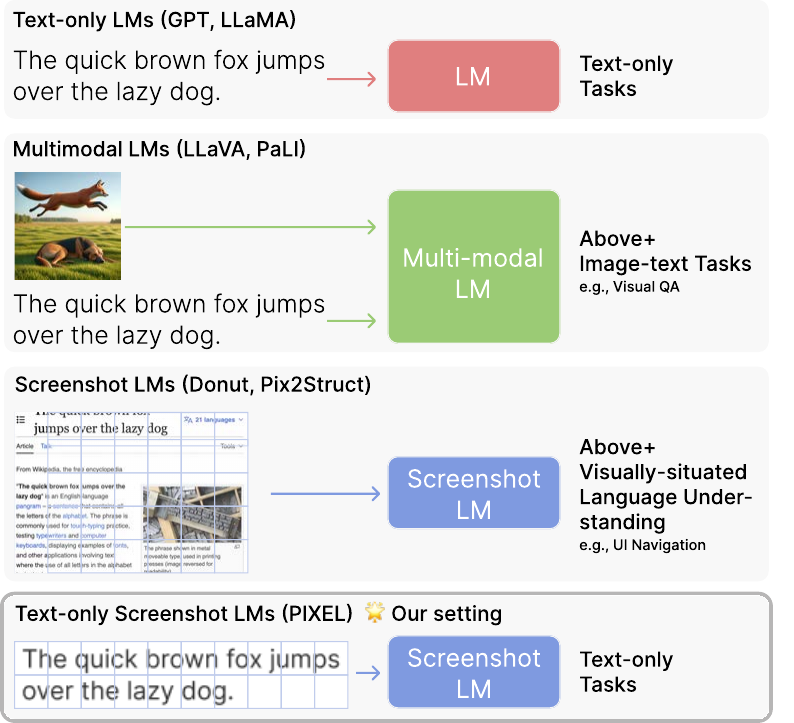}
    \caption{Illustrations of different LM paradigms and their applications. 
    We focus on improving the text understanding ability of screenshot LMs,
    and adopt a text-only screenshot setting for direct comparisons.
    }
    \label{fig:teaser}
\end{figure}

The capability for language models (LMs) to process both text and images in a single visual input
will enable a broad range of complex applications, 
such as 
document understanding~\cite{mathew2021docvqa},
chart reading~\cite{masry-etal-2022-chartqa}, and
UI navigation~\citep{zheran2018reinforcement}.
Previous methods relying on an off-the-shelf OCR pipeline \cite{huang2022layoutlmv3} are prone to error propagation~\citep{kim2022ocr}; 
multimodal models that process images and text separately~\cite[][\emph{inter alia}]{alayrac2022flamingo,liu2023llava} 
lose the spacial information between different elements.
Even frontier LMs such as GPT-4V~\citep{openai2023gpt4} and Gemini~\citep{team2023gemini} struggle 
at recognizing and understanding text within images~\citep{qi2023gemini}.

\paragraph{Screenshot LMs.}
A new family of models~\citep{lee2023pix2struct,rust2023language} has emerged that 
processes text---along with images, charts, and tables---all through one visual input.
We refer to them as \textit{screenshot} LMs.
They can be trained and deployed over vast amounts of ``screenshots'' available on the internet
such as webpage screenshots, UI images, and scanned documents.
They are designed to 
handle visually-situated text in
an end-to-end manner, 
and have the potential to unlock 
many powerful applications,
as illustrated in \Cref{fig:teaser}.

\paragraph{Challenges in understanding text from screenshots.} %
Recent developments
in screenshot LMs %
have shown promising results in 
certain scenarios, such as 
PIXEL~\cite{rust2023language} in multilingual transfer,
PhD~\cite{borenstein-etal-2023-phd} in historical document understanding, 
and Pix2Struct~\cite{lee2023pix2struct} in chart and UI understanding. %
However,
the mismatch in modality makes it challenging for screenshot LMs to 
effectively process the text in the inputs, and 
they exhibit a noticeable shortfall %
in language understanding tasks 
when compared to text-only LMs: 
For example, the prior state-of-the-art, PIXEL, still has a 7\% performance gap on GLUE~\citep{wang2019glue}
when compared to BERT~\citep{devlin2019bert}.
This disparity significantly restricts the utility of screenshot LMs for widespread applications. 
To integrate screenshot LMs in practical, real-world scenarios effectively, 
it is crucial to \textit{first} close this gap on text-only tasks.

\paragraph{Text-only screenshot LMs.}
To enhance the language understanding capability of screenshot LMs,
we focus on a \textit{text-only} screenshot setting, where 
inputs consist exclusively of plain text rendered as images. %
This particular setting facilitates direct comparison with text-only LMs, 
enabling us to isolate the impact of the quality of the screenshot data and 
focus on architecture and training objective changes to improve language understanding.

\begin{figure}[t]
    \center
    \includegraphics[width=0.90\linewidth]{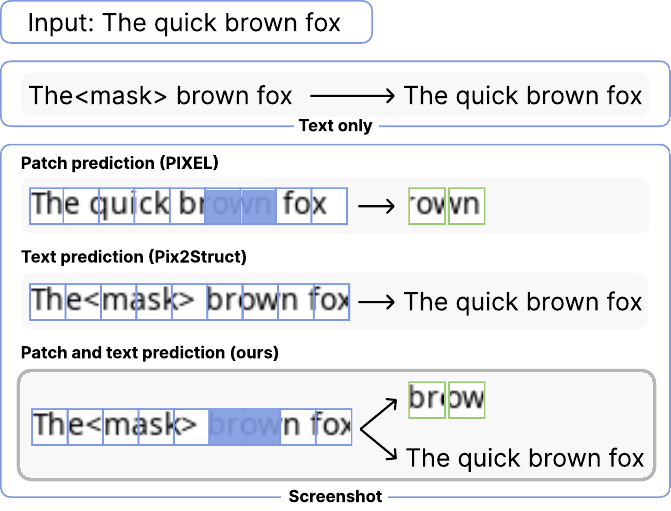}
    \caption{A comparison between existing objectives (PIXEL, Pix2Struct) and our \ours{} objective for training screenshot LMs.
    The blue grids illustrate how input images are split into patches.%
    }
    \label{fig:comparison}
\end{figure}

\paragraph{Our contributions.}
(1) 
We introduce the
\textbf{P}atch-and-\textbf{T}ext \textbf{P}rediction (\ours{}) objective. 
As shown in \Cref{fig:comparison},
previous works either only predict image patches or only predict text.
Instead, we mask and predict both the screenshot patches and text within the screenshot.
The choice is backed up by the intuition that
a patch prediction objective helps learn local visual features of the text, %
while a text prediction objective is more effective in learning to understand the language. %

(2) 
We find that screenshot LMs
often exhibit training instability
and are sensitive to hyperparameter choices.
We conduct careful ablations on 
masking rates and patch sizes, 
as well as exploring designs to stabilize the training of screenshot LMs.
Our pre-trained screenshot LM %
demonstrates strong language understanding capabilities:
it achieves comparable performance (within 2\%) to BERT$_\text{base}$~\cite{devlin2019bert} on 6 out of 8 datasets from GLUE~\cite{wang2019glue},
improving over previous best by up to \improve{}\%. 

(3) We  extend our objective to autoregressive LMs,
by feeding image patches and text tokens to a single decoder and predicting both modalities in an autoregressive manner.
We show that the autoregressive screenshot LMs can effectively utilize the screenshot context 
to reduce the perplexity on the subsequent text,
either by training from scratch or fine-tuning from existing text-only LMs.

We also discuss the limitations of current screenshot LMs, 
as well as possible future directions.
We hope our findings will inspire more research exploring screenshot LMs and their applications.

\begin{figure*}[t]
    \center
    \includegraphics[width=0.95\textwidth]{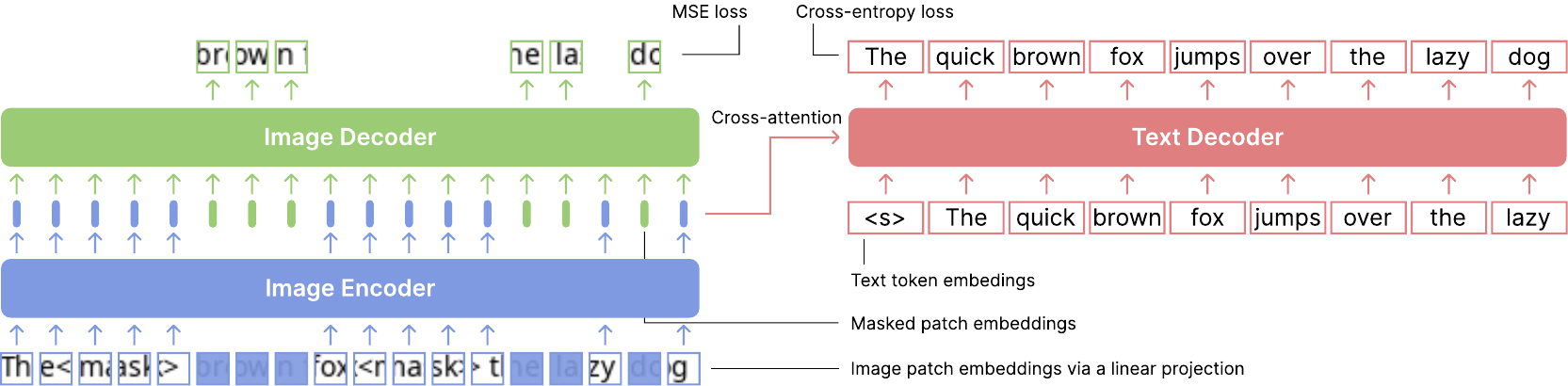}
    \caption{An illustration of our 
    \tf{P}atch-and-\tf{T}ext \tf{P}rediction
    (\ours) 
    objective. \ours{} applies both image-patch masking (dark blue grids) and text masking (\ttt{<mask>} tokens) to the input. Subsequently, an image decoder is used to reconstruct the masked patches, and a text decoder is used to recover the corrupted text.
    The illustration does not reflect all implementation details (e.g., the CLS token and subwords). 
    \Cref{app:pre-training-hyper} provides more details.
    }
    \label{fig:bidirectional}
\end{figure*}

\section{Problem Setup} %
\label{sec:setting}

In this section, we define our \textit{text-only screenshot LM} setup. 
For pre-training,
we assume a text corpus $\mathcal{C}$
and we render each text sequence $s\in \mathcal{C}$
as an image $I_s$. 
The model is allowed to utilize both $s$ and $I_s$ for its pre-training.
For fine-tuning and evaluation on the downstream dataset $\mathcal{D}$=$\{(x_i,y_i)\},1\leq i\leq |\mathcal{D}|$,
we render each input sequence $x$ as $I_{x}$ and define
$\mathcal{D}_I=\{(I_{x}, y)\},(x,y)\in \mathcal{D}$.
Then we only fine-tune and test the model
on $\mathcal{D}_I$.
In other words, 
the model can leverage both the ground-truth text and the rendered images 
for pre-training,
but can only take the rendered images as input for downstream tasks.
This is similar to a realistic end-to-end screenshot LM scenario,
where the inputs are predominately screenshots 
during deployment.

\section{\ours{}: Patch and Text Prediction} %
\label{sec:method_bi}

We introduce 
our screenshot LMs and our training objective \ours{}.
All the components of our model
are Transformers~\citep{vaswani2017attention}
or Vision Transformers (ViT; \citealp{dosovitskiy2021an})
and the architecture details can be found in \Cref{app:architecture}.
In the following,
we first introduce how the input images are processed, and 
then describe our rendering strategy and training objectives.

\subsection{Input Processing}
\label{sec:method_bi:input}

The input image (screenshot, in our case) is split into $n$ non-overlap patches, each sized $p_h \times p_w$, where $p_h$ and $p_w$ denote the patch height and width, respectively. By default, we use $p_h=p_w=16$, a setting commonly adopted in ViTs~\citep{dosovitskiy2021an}.
Each patch %
can be seen as a vector input $x_i \in \mathbb{R}^{p_h \times p_w \times c}$, where $i \in \{1, ..., n\}$ and $c$ is the number of channels (by default $c=3$). These patches are turned into patch embeddings via a linear layer, and are then fed as input features to the Transformer. 
ViTs use a fixed sine/cosine 2D positional embedding~\citep{vaswani2017attention}.

For the autoregressive text decoder,
we use the same tokenization as RoBERTa~\citep{liu2019roberta} and OPT~\citep{zhang2022opt}.

\subsection{Rendering Screenshots}
\label{sec:method_bi:rendering}

We mostly follow PIXEL \citep{rust2023language} for its rendering strategy: 
we render the text into an image of size $p_h \times np_w$, where $n$ is the number of patches. 
We replace all the new-line characters with a special symbol  and render the text into one line. 
We use the Google \texttt{Noto Sans} fonts collection\footnote{\url{https://fonts.google.com/noto}}.

We implement our own rendering engine, described in \Cref{app:rendering}.
By default, we use a font size of 10px, 
similar to the one used in PIXEL. 
On average, one $16\times 16$ patch can fit 
$0.57$
OPT text token. In other words, to encode the same amount of text,
a screenshot LM 
following the above rendering strategy
will require almost twice the number of tokens compared to a text-only LM.

\subsection{Training Objectives}
\label{sec:method_bi:objective}

We adopt two training objectives: 
a masked patch prediction objective and a masked text prediction objective. 
While either objective has been explored in prior work,
we are the first to combine them 
and show that it is critical to combine both objectives
to achieve competitive performance for screenshot LMs.
The intuition is that
patch prediction helps learn the local visual features
while text prediction is more effective for language understanding.
\Cref{fig:bidirectional} illustrates our training objective.

\paragraph{Patch masking and prediction.}
We randomly exclude input patches from the image encoder and 
use a bidirectional image decoder to recover them. 
An MSE loss is calculated between the predicted and the target pixel values of the masked patches. 
This paradigm was first proposed in MAE~\citep{he2021masked} for natural images
and later adopted in PIXEL~\citep{rust2023language} for text-rendered images.

\begin{figure}[t]
    \centering
    \includegraphics[width=0.98\linewidth]{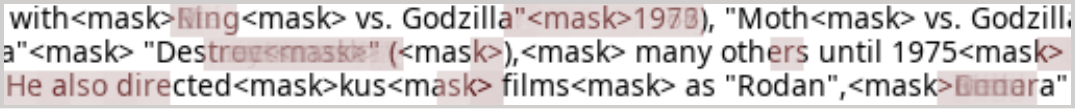}
    \caption{An example of the rendered text, the patch masks (red background; 25\% with span masking), and the prediction. 
    The image is one line of text but is
    cut and concatenated for better visualization here.
    }
    \label{fig:example_blur}
    \vspace{-10pt}
\end{figure}

In addition, we follow~\citet{rust2023language} and leverage their span masking strategy. %
As shown in Figure~\ref{fig:example_blur},
a single patch mask often leaves out shallow cues
where the model can  complete the word without learning the semantics.

We adopt a masking rate of \maskrate{}, which is drastically different from
75\% used in MAE and 25\% used in PIXEL.
In \Cref{sec:ablation_mask},
we demonstrate that when combined with the text masking objective,
it is important to retain the image masking objective
but best to keep a lower masking rate.

\paragraph{Text masking and prediction.}
The image prediction objective is effective in training the model to 
understand the visual representation of the text. 
However, they have several drawbacks: 
(1) the training procedure is often unstable due to the extreme contrast of black and white pixels, especially when the masking rate is high;
(2) patch masking, even with the span masking strategy, still often leaks shallow cues;
(3) pixel prediction does not model uncertainty well. While a text objective can model a probability distribution over a vocabulary, 
the image decoder often predicts blurry gray pixels, or 
a superposition of several possible words, as shown in \Cref{fig:example_blur}.

We believe that a text prediction objective 
is more effective in learning the language,
and combining it with the patch prediction 
can lead to both strong visual and text understanding capabilities.
We first randomly mask out tokens in the input text 
and then render the corrupted text as the screenshot. 
We then add an autoregressive text decoder 
to recover the corrupted text, similar to BART~\citep{lewis-etal-2020-bart}.
By default, we use a masking rate of 25\% with uniform masking, 
replacing masked tokens with a special token \texttt{<mask>},
and merging adjacent \texttt{<mask>} tokens.

\subsection{Designs to Stabilize Training}
\label{sec:training_stability}

In our preliminary experiments,
we observe that training screenshot LMs is often unstable. 
This is likely caused by %
the highly polarized and imbalanced distribution of pixel values in text-rendered images (either black or white, and mostly white).
We identify several design choices that
are critical,
removing of which can lead to performance degradation, training stagnation, or loss spikes. 

\textbf{Pixel-value preprocessing.}
We follow \citet{rust2023language} and adopt the following preprocessing: (1) The input pixel values are normalized to $[0, 1]$;\footnote{The other common 
strategy is to standardize the input pixel values, 
which is adopted by \citet{lee2023pix2struct}. However, we find such preprocessing hinders the training stability and the performance in the text-only screenshot setting.} 
(2) The ground-truth pixel values used for calculating the MSE loss 
are standardized by the means and standard deviations of the \textit{patch}, \textit{i.e.,} $\text{target}'=(\text{target}-\text{mean}) / \text{std}$.
While the input normalization is critical for training stability,
we find the target standardization significantly improves the final performance.

\textbf{Attention masks and end-of-sequence patch.}
Following~\citet{rust2023language},
we mask out attentions to the white empty patches after the end of the text sequence;
we also add an end-of-sequence black patch at the end of the text.
Removing either  will lead to higher chances of training collapse.

\textbf{Text prefix.}
We observe that the patch prediction loss of screenshot LMs
often goes through 
a ``plateau'' phase at the beginning of training,
as shown in \Cref{fig:maeloss}.
In this phase, the model only learns to predict grey blur patches
and with a certain chance, the training loss  never decreases and the training stagnates.
We find that rendering a text prefix at the beginning of the input sequence can 
reduce the chance of stagnation and start the loss decrease earlier. 
The intuition is that the text prefix  provides the model with an easy and stable target to learn. 
We also observe that a longer text prefix has more significant effect.
In our experiments, all models have a text prefix of ``\ttt{Beginning of the sequence:}''.
\Cref{app:example} shows examples of the rendered screenshots with the text prefix.

\textbf{Embedding layernorm.}
We observe that a model with a higher masking rate has a much higher chance to suffer from training collapse.
In such cases, adding a layer normalization  immediately after the input embeddings can  mitigate this problem while not inducing much change in the absolute loss value.
Though our main model does not have the embedding layernorm, 
some of our ablations employ the design to accommodate training stability.
We offer more details in \Cref{app:moreresults}.

\subsection{Fine-tuning on Downstream Tasks}
\label{sec:method_bi:finetune}

We examine our screenshot models
in a similar way as 
BERT: %
we fine-tune and evaluate them on natural language understanding tasks, e.g., GLUE. 
Since our model has an encoder-decoder architecture, %
we consider two possible solutions for fine-tuning:

\paragraph{Encoder-only.}
We simply keep the image encoder and discard both the image and the text decoders. 
Following PIXEL,
we take the average of the last layer representation from the image encoder as %
the sentence representation,
and feed it to a linear layer for classification or regression tasks.
By default, we use the encoder-only fine-tuning for evaluation.

\paragraph{Sequence-to-sequence (s2s).}
We  leverage the combination of the image encoder and the text decoder for downstream tasks via fine-tuning. 
The model is trained to directly generate text for these tasks (\textit{i.e.,} ``good'' or ``bad'' for a sentiment classification task). For all the details, please refer to \Cref{app:ft}.

\begin{table*}[t]
    \centering
    \resizebox{0.98\textwidth}{!}{
        \begin{tabular}{lrccrcccccccc}
            \hline
            \toprule
            \tf{Model}& $|\theta|$& \tf{PP} & \tf{TP} & \tf{MNLI} & \tf{QQP} & \tf{QNLI} & \tf{SST-2} & \tf{CoLA} & \tf{STS-B} & \tf{MRPC} & \tf{RTE} & \tf{Avg.} \\
            \midrule
 BERT$^\dagger$& 110M &  - & -  & 84.0/84.2& 87.6& 91.0& 92.6& 60.3& 88.8& 90.2& 69.5& 83.0\\
 \midrule
 DONUT$^*$ &  143M & \xmark & \cmark & 64.0/- & 77.8 & 69.7 & 82.1 & 13.9 & 14.4 & 81.7 & 54.0 & 57.2\\ 
 CLIPPO$^\clubsuit$ & 93M& \xmark & \xmark & 77.7/77.2 & 85.3 & 83.1 & 90.9 & 28.2 & 83.4 & 84.5 & 59.2 & 74.0\\
 PIXEL$^\dagger$& 86M&  \cmark & \xmark &78.1/78.9& 84.5& 87.8& 89.6& 38.4& 81.1& 88.2& 60.5& 76.0\\
 PIXAR$^\diamond$ & 85M & \cmark& \xmark & 78.4/78.6 & 85.6 & 85.7 & 89.0 & 39.9 & 81.7 & 83.3 & 58.5& 75.3\\ 
\bsl \colorcello {\bi{}} & \colorcello 86M& \colorcello \cmark & \colorcello \cmark & \colorcello \tf{80.9}/\tf{81.1} & \colorcello \tf{87.4}& \colorcello \tf{89.6} &\colorcello \tf{92.0} & \colorcello \tf{45.7} & \colorcello \tf{87.2} &\colorcello \tf{89.7} &\colorcello \tf{68.7} &\colorcello \tf{80.2} \\ %
\midrule
\bsl \colorcello {\bi{}}$_\text{s2s}$ & \colorcello 268M & \colorcello \cmark & \colorcello \cmark & \colorcello 82.2/82.6 & \colorcello 87.7 & \colorcello 90.4 & \colorcello 92.5 & \colorcello 48.8 & \colorcello 83.8  & \colorcello 90.6 & \colorcello 67.7 & \colorcello 80.5\\
        \bottomrule
        \end{tabular}
    }
    \caption{
        Validation results for \bi{} and baseline models fine-tuned on GLUE. 
        We report F1 scores for QQP and MRPC, Matthew's correlation for CoLA, Spearman's correlation for STS-B, and accuracy for others. 
        We report baseline results from \citet{rust2023language}$^\dagger$, \citet{borenstein-etal-2023-phd}$^*$, \citet{tschannen2023clippo}$^\clubsuit$, and \citet{tai2024pixar}$^\diamond$.
        The averaged results are calculated without MNLI (mismatched).
        We report the number of parameters $|\theta|$ used in fine-tuning.
        PP: the model uses patch prediction; TP: the model uses text prediction.
        ``\ours{}'' denotes the encoder-only fine-tuning and
        ``\ours{}$_\text{s2s}$'' denotes the sequence-to-sequence fine-tuning.\protect\footnotemark{}
    }
    \label{tab:main_glue}
\end{table*}

\section{Experiments}
\label{sec:exp}

\subsection{Setup}
\label{sec:exp_setup}

\paragraph{Pre-training.} 
We pre-train our models on the English Wikipedia and BookCorpus~\citep{zhu2015aligning} corpora for 16 epochs with a batch size of 256 (roughly 1M steps).
For each input instance, we render a 256-token text sequence to an image with size
$16\times 8192$ (512 patches). %
More details on pre-training 
are provided in \Cref{app:pre-training-hyper}.

\paragraph{Fine-tuning.}
We fine-tune our models on datasets from the GLUE benchmark~\citep{wang2019glue}. 
We run grid search and report the average validation performance of 3 random seeds.
\Cref{app:ft} provides more details on fine-tuning.

\paragraph{Baselines.}
We compare our models with several text-only and screenshot LM baselines.

\textbf{BERT}~\citep{devlin2019bert}. We compare to BERT because it is a bidirectional text-only LM at a similar scale (110M) and it uses the same pre-training corpora as ours. 
Note that BERT 
adopts more epochs over the training data than ours (40 epochs vs. our 16 epochs).

\textbf{Donut}~\citep{kim2022ocr}. 
Donut is an encoder-decoder model that takes document images as input 
and outputs text. 
It is pre-trained with a pseudo-OCR task on IIT-CDIP~\citep{lewis2006building} %
and synthetic images generated from English, Chinese, Korean, and Japanese Wikipedia.

\footnotetext{
    For Pix2Struct, there are no publicly reported GLUE results. In our preliminary experiments, the performance on GLUE tasks significantly lags behind other baselines, 
    probably due to that it is not designed for text-only tasks. 
    Thus we leave it out from our comparison.
}

\textbf{CLIPPO}~\citep{tschannen2023clippo}. 
This is a variant of CLIP~\citep{radford2021learning} that 
utilizes one single vision encoder to process both images and text that is rendered as images.
We report the GLUE performance of CLIPPO trained on 
WebLI~\citep{chen2023pali}
and 25\% C4~\citep{raffel2020exploring},
which is a \emph{significantly} larger pre-training corpus than ours.

\textbf{PIXEL}~\citep{rust2023language}. PIXEL is a screenshot LM that is trained with only a patch masking and  prediction objective. 
The other configurations (e.g., pre-training corpora and hyperparameters) are mostly the same as ours.
A comparison between PIXEL and our model can help us understand the effect of our new training objective.

\textbf{PIXAR}~\citep{tai2024pixar}. 
PIXAR is a concurrent work that explores autoregressive extensions of PIXEL. To be able to generate text via generating images, PIXAR adopts adversarial training and uses OCR softwares to extract the text. 
PIXAR uses the same training data as PIXEL.

\subsection{Main Results: \ours{} Outperforms Other Screenshot LMs Significantly}
\label{sec:exp_main_results}

Table~\ref{tab:main_glue} shows 
the main results on the validation sets of the GLUE benchmark. 
We also report the full results with standard deviation
in \Cref{tab:main_glue_rep_error} and the
test results of our models and reproduced baselines in \Cref{tab:main_glue_test}. %
Firstly, compared to previous  state-of-the-art
screenshot LMs,
our \bi{} achieves significantly better performance on almost all GLUE tasks,
indicating that our objective leads to better language understanding capabilities.
Our model improves upon the previous state-of-the-art 
by up to \improve{}\% on specific tasks and more than 4\% on average.
Comparing \bi{} to BERT,
the performance gap is substantially reduced---on 6 out of the 8 tasks, the difference is within 2\%
(for the previous best model, this number is 1 out of the 8 tasks).

\subsection{Ablation on Training Objectives}
\label{sec:ablation_obj}

Several previous works explored 
patch masking and text masking
separately~\citep{rust2023language,lee2023pix2struct},
while we combine both for the first time.
Here we conduct ablations to study the efficacy of 
(1) patch masking, (2) patch prediction, (3) text masking, and (4) text prediction.
Table~\ref{tab:ablation_obj} demonstrates a clear comparison:
while using the text objective alone is better than predicting image patches,
combining both leads to significantly improved results, 
showing that the model trained with both objectives 
can better understand the text within the screenshot images.

\begin{table}[t]
    \centering
    \resizebox{0.98\linewidth}{!}{
        \begin{tabular}{cccccccc}
            \hline
            \toprule
            \tf{PM} & \tf{PP} & \tf{TM}& \tf{TP} & \textbf{MNLI} & \textbf{SST-2} & \textbf{MRPC} & \textbf{RTE} \\
            \midrule 
            \cmark & \cmark  & \xmark & \xmark & 78.6 & 89.2 & 88.5  & 66.5 \\ %
            \cmark & \xmark  & \xmark & \cmark & 79.5 & 90.0 & 88.7 & 66.7\\
            \xmark & \xmark  & \cmark & \cmark & \tf{82.4} & 90.9 & 86.5 & 62.0\\ %
            \cmark & \xmark  & \cmark & \cmark & 81.1 & 91.0 & 88.1 & 63.8\\
            \colorcello \cmark & \colorcello \cmark &  \colorcello \cmark& \colorcello \cmark & \colorcello 80.9 & \colorcello \tf{92.0} & \colorcello \tf{89.7} & \colorcello \tf{68.7}  \\ %
        \bottomrule
        \hline
        \end{tabular}
    }
    \caption{
        Ablations on different training objectives.  
        PM: patch masking; PP: patch prediction; TM: text masking; TP: text  prediction.
        The patch masking rate is 10\% with span masking and the text masking rate is 25\%.
    }
    \label{tab:ablation_obj}
\end{table}

\subsection{Ablation on Masking Rates}
\label{sec:ablation_mask}

For masking-style pre-training (like masked language modeling or masked auto-encoding in vision),
the masking rate is one of the most important hyperparameters.
\citet{he2021masked} show that for natural images, 
the masking rate can be as high as 75\%; 
\citet{wettig-etal-2023-mask} demonstrate that for text,
masking rates can also be higher than the conventionally used 15\%~\citep{devlin2019bert},
but too high of a masking rate (e.g., larger than 50\%) will hurt the performance. 
In our case, we have two masking rates and 
we run a thorough grid search to find the best configurations.

Table~\ref{tab:ablation_mask} shows 
the comparison with different masking rates.
The first unique phenomenon
we observe is that 
a high patch masking rate (e.g., 40\%)
leads to frequent loss spikes and training collapses.
We then see that with a smaller patch masking rate (10\%) and 
a 25\% text masking rate,
the model can achieve the best fine-tuning performance.
We hypothesize that the patch masking mostly
helps learn the visual representations of the text (hence no need for a very high masking rate),
while the text masking and prediction objective plays a pivotal role in facilitating learning the language.
We also show in \Cref{tab:ablation_spanmasking}  that using \emph{span} patch masking improves over uniform patch masking.

\begin{table}[t]
    \centering
    \resizebox{0.87\linewidth}{!}{
        \begin{tabular}{cccccc}
            \hline
            \toprule
            \multicolumn{2}{c}{\textbf{Mask Rate}} & \multicolumn{1}{c}{\multirow{2}{*}{\textbf{MNLI}}} & \multicolumn{1}{c}{\multirow{2}{*}{\textbf{SST-2}}} & \multicolumn{1}{c}{\multirow{2}{*}{\textbf{MRPC}}} & \multicolumn{1}{c}{\multirow{2}{*}{\textbf{RTE}}} \\
            \cmidrule{0-1}
\multicolumn{1}{c}{Patch} & \multicolumn{1}{c}{Text} &  \multicolumn{1}{c}{} & \multicolumn{1}{c}{} & \multicolumn{1}{c}{} & \multicolumn{1}{c}{} \\
            \midrule 
            5\% & 25\% & 79.2 & 90.0 & 85.7 & 62.3 \\
            10\% & 10\% & 80.0 & 90.2 & 88.7 & 64.5\\
            \colorcello 10\% & \colorcello 25\% & \colorcello \tf{80.9} & \colorcello \tf{92.0} & \colorcello \tf{89.7} & \colorcello \tf{68.7} \\
            25\% & 25\% & 80.8 & 90.7 & 88.6 & 65.3\\
            10\% & 40\% & 79.6 & 90.2 & 85.1 & 62.2\\
            25\% & 40\% & 80.6 & 90.9 & 88.9 & 67.9 \\
        \bottomrule
        \hline
        \end{tabular}
    }
    \caption{
        Results on different masking rates.
        All patch masking has span masking.
    }
    \label{tab:ablation_mask}
\end{table}

\begin{table}[t]
    \centering
    \resizebox{0.98\linewidth}{!}{
        \begin{tabular}{cccccc}
            \hline
            \toprule
            \tf{PM} & \tf{Patch Size} & \tf{MNLI} & \tf{SST-2} & \tf{MRPC} & \tf{RTE} \\
            \midrule
            \multirow{2}{*}{25\%} & $16\times 16$ & 80.8 & 90.7 & 88.6 & 65.3\\ %
            \phantom{} & $16\times 32$ & 80.7 & {91.6} & \tf{89.7} & {67.4}\\ %
            \midrule
            \multirow{3}{*}{10\%} & \colorcello $16\times 16$ & \colorcello \tf{80.9} & \colorcello \tf{92.0} & \colorcello \tf{89.7} & \colorcello \tf{68.7}\\% 10span/25
            \phantom{} & $16\times 32$ & 80.4 & 91.2 &89.4 & 66.5\\% 10span/25 
            \phantom{} & $16\times 64$ & 76.7 & 88.5 & 88.1 & 64.4\\
            \bottomrule
            \hline
        \end{tabular}
    }
    \caption{
        Ablation study on patch sizes. 
        We show that 
        the optimal patch size depends on the masking rates. ``PM'' refers to the span patch masking rate of the input image. 
        All models here use a 25\% text masking rate.
    }
    \label{tab:ablation_patch}
\end{table}
\subsection{Ablation on Patch Sizes}
\label{sec:ablation_patch}

Table~\ref{tab:ablation_patch} shows that the best patch size varies depending on the masking rates:
when using a patch masking rate of 25\% (span), 
a larger patch size of $16\times 32$ is better;
when using a patch masking rate of 10\% (span),
the smaller $16\times 16$ is better;
using larger patches (e.g., $16\times 64$) leads to significant but non-catastrophic performance drops, yet they also come with efficiency gains by reducing the number of tokens needed to encode the same text content.
We hypothesize that patch sizes have a twofold effect:
while a larger patch size reveals less superficial clues, %
it also reduces the sequence length and leaves the model less compute to process the input. 
Note that one factor we ignore is that changing the patch size modifies the span masking behavior---larger patch sizes are effectively {larger spans}---and we leave it out for future research.

We draw a connection between this ablation study and changing the font size---increasing the patch size has a similar effect to decreasing the font size, as both squeeze more content into one patch. 
We alternate the patch size instead of the font since the patch size can be more intuitively controlled. 

A related topic is patch alignment: 
\citet{lotz-etal-2023-text} study rendering strategies for PIXEL where 
they align words and patches such that, for example, every two patches correspond to one word.
This strategy helps improve the downstream task performance,
but we argue that considering the potential usage of screenshot LMs (webpages and scanned documents),
such alignment is hard to achieve in practice and we leave it out of our ablations.

\section{Extension to Autoregressive ~ Screenshot LMs}
\label{sec:auto}

\begin{figure}[t]
    \center
    \includegraphics[width=0.98\linewidth]{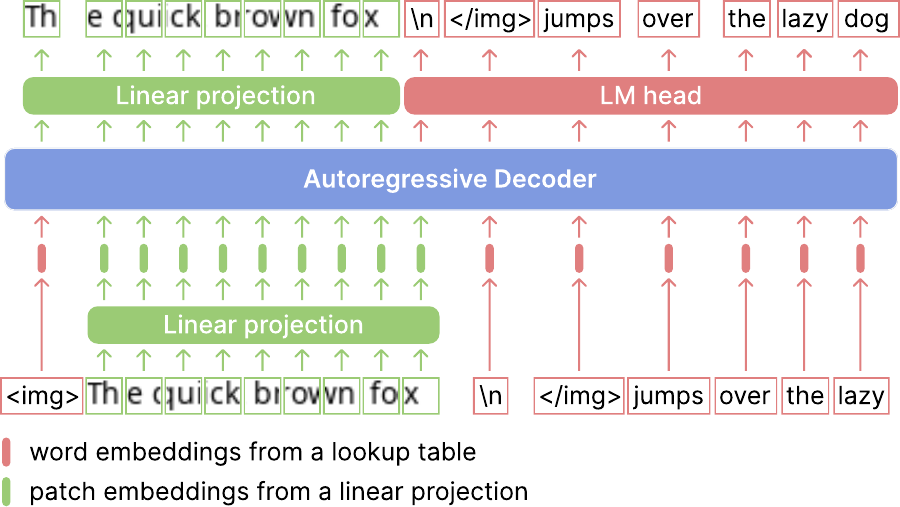}
    \caption{Our autoregressive screenshot LM.  %
    }
    \label{fig:auto}
\end{figure}

Autoregressive LMs have become the predominate form of large LMs
due to their %
powerful generation capabilities 
and emerging properties such as in-context learning \citep{brown2020language,openai2023gpt4,touvron2023llama}.
Nevertheless,
existing screenshot LMs\footnote{Concurrent work PIXAR~\citep{tai2024pixar} also explores autoregressive screenshot LMs but they generate images and rely on OCR softwares to convert them into text. } are 
built with 
an encoder-decoder architecture.
In this section, we explore the feasibility of 
extending \ours{} to 
decoder-only, autoregressive screenshot LMs.

\subsection{Methods}
\label{sec:auto_modeling}

Inspired by \citet{fuyu-8b},
we employ a single decoder architecture, 
where both visual and textual inputs are 
first mapped into token embeddings
and then processed by the same Transformer backbone.
The input of the model encompasses two segments: 
the first segment is a sequence of patches that collectively constitutes a screenshot, 
while the second segment is a sequence of text tokens that follows the screenshot content.
Unlike \citet{fuyu-8b} which only train the model to predict on the text segment, 
we apply an autoregressive objective on \emph{both} the screenshot and the text segments. 
\Cref{fig:auto} illustrates our autoregressive model and 
details are as follows.

\paragraph{Input format.}
Given a text sequence with $m$ text tokens, we first split it 
into the screenshot segment ($m_s$ tokens)
and the text segment ($m_t$ tokens), where $m_s + m_t = m$.
The screenshot image is of size $p_h \times np_w$, 
where $n$ is the number of patches. %
We insert three special tokens into the screenshot segment: 
a beginning-of-image token \texttt{<img>}, 
an image new line token \texttt{\textbackslash n}\footnote{This is to follow \citet{fuyu-8b} 
and allow the model to know that there is a new line of patches without using two-dimensional positional embedding. 
}, 
and
an end-of-image token \texttt{</img>}.
The special tokens are added to inform the model the boundary between the screenshot and text segments. 
In our experiments, we set $m_s=m_t=256$ and $n = 512$.

By default, we use the same rendering strategies as our encoder-decoder model: we use the Google \ttt{Noto Sans} font and a font size of 10px.
On average, one text token from the LLaMA~\citep{touvron2023llama} tokenizer takes 1.28 image patches when the patch size is $16\times16$. 

\paragraph{Architecture.}
We follow the same architecture as LLaMA~\citep{touvron2023llama}, 
one of the most popular open-source autoregressive LMs. 
We experiment with two model sizes: (1) a smaller model with 380M parameters which we train from scratch, and (2) a larger model with 1.3B parameters which we continue training from a pre-trained Sheared-LLaMA checkpoint \citep{xia2023sheared}. 
For detailed configurations, please refer to \Cref{app:auto}.

For the image patch input, we use a linear projection %
to map the pixel values to patch embeddings; %
For the text token input, we use their corresponding word embeddings. 
The patch and text embeddings are jointly fed into the Transformer blocks. %

\paragraph{Training objectives.}
We adopt both a next  patch prediction and a next  token prediction objective, as illustrated in \Cref{fig:auto}. 
For the next patch prediction, we take the last layer representation, use a linear projection 
to map it to pixel values, and calculate the MSE loss. %
For the next text token prediction, 
we %
use an LM head and calculate the cross-entropy loss.

\subsection{Experiment Results}
\label{sec:auto_results}

\paragraph{Evaluation setting.}
Our main goal is to verify whether 
the autoregressive screenshot LMs can understand the text from the screenshot context.
The screenshot LM is given 256 text tokens in screenshot context
and 256 text tokens in the text format,
and its text-only counterpart is given just 256 text tokens.
Then we measure perplexity on the last 256 text tokens.
If the screenshot LM can effectively utilize the screenshot context,
it will achieve lower perplexity compared to the text-only baseline.

\paragraph{Training from scratch.}
We train 380M parameter LLaMA-based models from scratch on the English Wikipedia and BookCorpus~\citep{zhu2015aligning} corpora for 16 epochs.
Each training instance is composed of  512  patches (rendered from a sequence of 256 text tokens) 
and its subsequent 256 text tokens.
We also train a text baseline with the same configuration, except its inputs are sequences of 512 text tokens.
Details are in \Cref{app:auto}.

\begin{table}[t]
    \centering
    \resizebox{0.98\linewidth}{!}{
        \begin{tabular}{llr}
            \toprule
            \tf{Model} & \tf{Context} & \tf{PPL} \\
            \midrule
            Text only & None  & 10.28\\ 
            Ours  & 256 text tokens (in screenshots)   & 9.66\\ 
            \tableindent w/o patch pred  &256 text tokens (in screenshots) &  10.77 \\
            \midrule
            Text only & 256 text tokens (in text)    & 8.60\\ 
        \bottomrule
        \end{tabular}
    }
        \caption{
        Comparisons between 380M LMs trained from scratch. %
        The input consists of two parts: 
        a segment of additional ``context'', and subsequent 256 text tokens which the perplexity (PPL) is evaluated on.
        ``w/o patch pred'': without the patch prediction objective.
    }
    \label{tab:auto_scratch}
\end{table}

\begin{table}[t]
    \centering
    \resizebox{0.94\linewidth}{!}{
        \begin{tabular}{llr}
            \toprule
            \tf{Model} & \tf{Context} & \tf{PPL} \\
            \midrule
            Text only & None & 10.20 \\
            Ours & 256 text tokens (in screenshots) & 9.09\\
            \midrule
            Text only & 256 text tokens (in text) & 7.68\\
        \bottomrule
        \end{tabular}
    }
    \caption{
    Comparison between 1.3B LMs fine-tuned from Sheared-LLaMA~\citep{xia2023sheared}.
    }
    \label{tab:auto_llama}
\end{table}

Table~\ref{tab:auto_scratch} 
shows that our autoregressive screenshot LM 
is able to effectively utilize the screenshot context
and reduce the validation perplexity ($10.28\rightarrow9.66$). 
We also conduct an ablation {without the patch prediction objective}, %
which performs significantly worse.
When compared to the text-only baseline using the same additional context but in text modality,
there is still a significant gap.
We hypothesize that the gap comes from two aspects: 
(1) training on screenshots is less effective than training on plain text; 
(2) it is more challenging to process the content in screenshots than in text.

\paragraph{Fine-tuning pre-trained LMs.}
We  fine-tune a pre-existing text-only LM, 
Sheared-LLaMA-1.3B~\citep{xia2023sheared}, 
with our autoregressive screenshot objective on RedPajama
~\citep{together2023redpajama} for 5 billion tokens.
We also fine-tune a text-only version with the same data for comparison. 
More details are provided in \Cref{app:auto}.

Table~\ref{tab:auto_llama} shows that %
our objective
can be effectively deployed for fine-tuning an existing LM, where our model 
improves the perplexity by using the additional screenshot context.
Though %
the screenshot model still has a substantial gap to the text-only baseline when 
the text-only model uses the same context in text,
it is a first step 
towards effective autoregressive screenshot modeling.

\section{Related Work}
\label{sec:related_work}

\paragraph{Multimodal LMs.}
A majority of work along the line 
focuses on effective adaptation of visual representations---acquired via a separate visual encoder~\citep{radford2021learning,dosovitskiy2021an}---into the text LMs.
One approach is to incorporate the visual representations via cross-attention~\cite{alayrac2022flamingo, li2023otter,bai2023qwenvl}.
More works directly 
use visual embeddings
as input ``tokens'' of LMs~\cite{lu2019vilbert,liu2023llava, liu2023improvedllava, zhang2023llamaadapter, gao2023llamaadapterv2, wang2023cogvlm,chen2023pali,driess2023palme}, 
sometimes with additional processing~\citep{li2023blip2,instructblip,zhu2023minigpt}.
\citet{fuyu-8b,li2023otterhd} instead 
directly process the  patches with a linear layer and use the embeddings as input,
omitting the additional visual encoder.

\paragraph{Screenshot LMs.} 
Two motivations inspire the development of screenshot LMs in previous literature.
The first one is to develop tokenizer-free models 
for the purpose of 
better cross-lingual transferability. %
Early work dates back to \citet{meng2019glyce} which explore glyph embedding for Chinese characters;
\citet{salesky-etal-2021-robust} adopt visual text representations on machine translation tasks to improve robustness. 
The representative work along this line is PIXEL, %
where \citet{rust2023language} train a ViT-MAE model over text-rendered images 
with masked patch prediction. 
Compared to its text-only counterpart,
PIXEL achieves better performance on non-Latin languages and is more robust toward orthographic noises,
but it lags behind on English tasks. %
Subsequent literature explores rendering strategies~\citep{lotz-etal-2023-text},
extension to historical documents~\citep{borenstein-etal-2023-phd}, 
and generating text via generating images~\citep{tai2024pixar,li2023renderdiffusion}.

The second motivation is to build end-to-end systems for understanding visually-situated text, for example, 
scanned documents, webpages, or UIs.
Early works are mostly pipeline systems where they use OCR tools~\citep{huang2022layoutlmv3,powalski2021going,appalaraju2021docformer} or
feed view hierarchy information~\cite{li-etal-2020-mapping,baiuibert}.
Recent works explore end-to-end models that take in 
solely visual inputs and generate text~\citep{li2023spotlight,davis2022end,aggarwal-etal-2023-dublin,kim2022ocr,zhu2023efficient}.
Pix2Struct~\citep{lee2023pix2struct}, as one of the latest development, 
pre-trains encoder-decoder models on large-scale webpage screenshots 
by masking out certain HTML elements and predicting the masked HTML code.
It achieves state-of-the-art performance on several visually-situated language understanding tasks and
inspires followup works on table understanding~\citep{alonso2023pixt3}, UIs~\citep{shaw2023from}, and more.

Though both approaches build promising systems for specific applications,
their basic 
text understanding capabilities still lag far behind text-only LMs,
which limits their deployment in general domains---and this is the gap we aim to close.

\section{Conclusion}

We introduce \ours{}, %
a new objective for training screenshot LMs by
predicting both masked image patches and masked text.
Our model
significantly improves the performance of screenshot LMs 
on GLUE and 
pushes the language understanding performance of screenshot 
LMs closer to their text-only counterparts.
We also demonstrate the effectiveness of \ours{} %
on autoregressive screenshot LMs. %

Numerous challenges still remain in the development of screenshot LMs.
For example, 
the patch prediction objective often makes the training unstable; 
the model is less efficient than the text-only LMs due to 
the longer input. 
Improving the stability and the efficiency of screenshot LMs,
as well as integrating real-world screenshots such as 
webpages into the training data, are promising future directions.
We hope that our work will inspire more effort in the direction, 
enabling more powerful screenshot LMs in the future.

\section*{Limitations}

Our work focuses on text-only screenshot LMs,
which is a simplified setting of general screenshot LMs.
Though we argue that 
understanding text  is the most challenging and fundamental aspect of screenshot LMs, 
we acknowledge that  our findings may not generalize to
the real screenshot scenarios, which we will explore in future work.

Our ablation study, though extensive, is not exhaustive.
For example, due to limited computational resources,
we are unable to explore all possible combinations of masking rates and masking strategies.
Considering that changing the patch size will also affect the masking (larger patches will essentially lead to more span masking),
a more comprehensive study is needed to fully understand the effect of masking and the optimal setting.
We also did not thoroughly explore the pre-training hyperparameters and followed existing works.
Regardless, 
we believe that the above limitations do not affect our main findings and contributions.

\section*{Acknowledgements}

We appreciate feedback from the members of the 
Princeton NLP group and Princeton Language and Intelligence.
We thank
Sadhika Malladi,
Alexander Wettig, Mengzhou Xia, Chenglei Si,
Alexis Chevalier,
Saumya Malik,
Elizabeth Salesky, Jonas F. Lotz,
Kenton Lee, and Mandar Joshi
for the valuable discussions and suggestions.
Tianyu Gao is supported by an IBM PhD Fellowship.
This work is also supported by a Sloan Research Fellowship.

\bibliography{custom}

\clearpage

\appendix

\onecolumn

\section{Rendering Strategy}
\label{app:rendering}

There are two rendering strategies: (1) pre-rendering the text and storing the screenshots, or (2) rendering the text on-the-fly during training.
While the first one is more efficient (for training), 
it requires a large amount of storage space
and has limited flexibility (e.g., we need to regenerate the whole dataset if we need to change the font).
We choose to render the text in an online manner,
and thus we need a fast renderer to avoid data processing becoming a bottleneck.

At the time of writing, various renderers are available that provide varying combinations of features. 
However, they are either too slow or not compatible with PyTorch's multi-process data loading. 
We then develop our own in-house renderer.
Our renderer is implemented in C++ and meshed to Python via PyBind11 \cite{pybind11}. 
We use the FreeType library to get the glyphs for the characters.
It is helpful that FreeType already provides the horizontal and vertical offsets to be applied to each character, so we can simply render each character in turn.
We allow the caller to control the font, font size, height, width, line spacing, word spacing, and margins. %
We observed roughly a $6.4\times$ speedup compared to PyGame\footnote{\url{https://github.com/pygame/pygame}}, a renderer used by \citet{rust2023language}.

\section{Model Architectures}
\label{app:architecture}

Our model architecture mainly consists of three components: 
(1) an image encoder; 
(2) an image decoder; 
and (3) a text decoder. 
All of these components are based on Transformers. 
Following \citet{rust2023language}, we add a \texttt{CLS} token at the beginning for the encoder input. 
During pre-training, the image encoder takes input from unmasked image patches. 
The encoder output is used by the image decoder (along with masked tokens, represented by a mask embedding).
The image decoder predicts only on the masked patches. 
The text decoder uses cross attention to attend to 
outputs of the image encoder (hence only the representations of unmasked image patches). 

We strictly follow \citet{rust2023language,he2021masked}
for the image encoder and the image decoder settings. 
They are standard ViTs with pre-layer normalizations.
For the text decoder, 
we follow \citet{lee2023pix2struct}, which use a GLU (Gated Linear Unit; \citealp{dauphin2017language}) version of MLP~\citep{shazeer2020glu}.
We also change the positional embedding of the text decoder to a learnable absolute positional embedding for simplicity.
For ``our text LM'' (mentioned below), we use the same configurations as our encoder-decoder model except that we also use GLUs for the text encoder.
Specific hyperparameter settings for the architecture are provided in Table \ref{tab:arch}.
We use an image size of $16\times 8192$.
For some ablation models,
we have used slightly different image sizes (e.g., $16\times 8464$ to be consistent with \citealt{rust2023language}),
but they are roughly in the same range.

In our preliminary experiments, we find that adding a layernorm after the input linear projection will lead to more stable training and fewer loss spikes, while not changing the final performance much.
We adopt the input layernorm for some of the ablation models demonstrated in \Cref{tab:layer_norm_table}.

\begin{table}[hb!]
\centering
\scalebox{0.92}{
\begin{tabular}{llll}
\hline
\toprule
\multicolumn{1}{l}{\multirow{-1}{*}{\textbf{Component}}} & \textbf{Image Encoder} & \textbf{Image Decoder} & \textbf{Text Decoder} \\

\midrule
Image patch size & 16 $\times$ 16 & - & - \\
Hidden size & 768 & 512 & 768\\
Intermediate size & 3072 & 2048 & 3072\\
\#Attention heads & 12 & 16 &12\\
\#Layers & 12 & 8 & 12\\

\bottomrule
\hline
\end{tabular}
}
\caption{{Architecture configurations of \ours{}.} }
\label{tab:arch}
\end{table}

\section{Pre-training  Details}
\label{app:pre-training-hyper}

We use Huggingface's Tranasformers package~\citep{wolf2020transformers} to perform all our pre-training and fine-tuning experiments.
We provide the data and optimization hyperparameters during the pre-training of our \ours{} model in Table \ref{tab:pretrain_hp}.
We use FlashAttention~\citep{dao2022flashattention} to speedup training.
We also have several special design choices: 
(1) We always render a black patch at the end of the text, indicating the end of the sequence (following PIXEL);
(2) we do not attend to the white patches after the end-of-sequence black patch (following PIXEL);
(3) we normalize the input pixel values and standardize the target pixel values in each patch before calculating the MSE loss (following PIXEL); 
(4) we always render a prefix text \ttt{Beginning of the sequence:} at the beginning, which serves as an ``anchor point'' and helps warmup the training (new compared to PIXEL).
We find that the above designs are crucial for making the training stable and
avoiding training stagnation or loss spikes.

\begin{table}[hb!]
\centering
\scalebox{0.92}{
\begin{tabular}{ll}
\hline
\toprule
\multicolumn{1}{l}{\multirow{-1}{*}{\textbf{Parameter}}} & \textbf{Value} \\

\midrule
\textbf{{Data}} \\
\quad Image size (height, width) & $(16, 8192)$ \\
\quad Image mode & RGB \\
\quad Font & Google Noto Sans\\%  (\texttt{GoNotoCurrent}) \\
\quad Font size & $10$ \\
\quad Line space & $6$ \\
\quad Newline symbol & //// \\ %
\quad Patch masking rate & $10\%$ \\
\quad Patch span masking & \texttt{true} \\
\quad Patch span masking max length & $6$ \\
\quad Patch span masking cumulative weights & $\{0.2, 0.4, 0.6, 0.8, 0.9, 1\}$ \\ 
\quad Text masking rate & $25\%$ \\
\quad Text masking token & \texttt{<mask>} \\
\quad Merge consecutive text masks & \texttt{true} \\

\textbf{{Optimization}} \\
\quad Learning rate & $1.5e^{-4}$ \\
\quad Minimum learning rate & $1.0e^{-5}$\\
\quad Warmup & $50$K steps\\
\quad Learning rate scheduler & Cosine decay \cite{loshchilov2017sgdr} \\
\quad  Batch size & $256$ \\
\quad Optimizer & AdamW \cite{loshchilov2018decoupled} \\
\quad Mixed precision training & \texttt{fp16} \\
\quad Number of epochs & $16$ (roughly 1M steps)\\

\bottomrule
\hline
\end{tabular}
}
\caption{
{Hyperparameters in \ours{} pre-training.} 
}
\label{tab:pretrain_hp}
\end{table}

\section{Fine-tuning Hyperparameters}
\label{app:ft}

We fine-tune our models on datasets from the GLUE benchmark~\citep{wang2019glue}, including 
SST-2~\cite{socher2013recursive_sst-2}, CoLA~\cite{warstadt2019neural_cola}, MNLI~\cite{williams2018broad_mnli}, QNLI~\cite{rajpurkar2016squad}, RTE~\cite{dagan2005pascal_rte1,bar2006second,giampiccolo2007third_rte3,bentivogli2009fifth_rte4}, MRPC~\cite{dolan2005automatically_mrpc}, QQP\footnote{\url{https://www.quora.com/q/quoradata/}} and STS-B~\cite{cer2017semeval_sts-b}. 
We did not include WNLI due to its abnormal data distribution issue noted on the GLUE website\footnote{\url{https://gluebenchmark.com/faq}}.
We run grid search for fine-tuning and report the average of the best validation results over three seeds.
\Cref{tab:ft_hp} shows the hyperparameters used for fine-tuning.
For rendering, we use the same rendering engine, font, and font size as in pre-training.
We also add the ending black patch and the prefix text to be consistent with pre-training.
We mask out the attention to the white patches after the  black patch.
We use an image size of $(16, 8192)$ for MNLI, QQP, QNLI, and RTE, and $(16, 4096)$ for the rest.
We evaluate the model every $100$ steps for MRPC, STS-B, and CoLA, every $250$ steps for RTE, and every $500$ steps for the remaining tasks.
Unlike text models like BERT,
we find that screenshot LMs are sensitive to the number of 
optimization steps instead of epochs of data, 
thus we control the total number of training steps, similar to \citet{rust2023language}.

For sentence pair tasks, we render a \ttt{////} between the two sentences. 
We replace all newlines in the text with \ttt{////}. 
For the sequence-to-sequence setting, 
we show the corresponding label text for each task in \Cref{tab:seq2seq_label}.
Specifically, for STS-B (a regression task), 
we follow the setting from T5~\citep{raffel2020exploring},
where we round up all the values to the nearest increment of 0.2; during evaluation,
we convert the output text to floats and compute the metric using the original label values.

\begin{table}[h!]
    \centering
    \scalebox{0.92}{
    \begin{tabular}{ll}
    \hline
    \toprule
    \tf{Parameter} & \tf{Value} \\
    \midrule
    Optimizer & AdamW \\
    Warmup steps & 100\\
    Learning rate scheduler & Linear decay \\
    Mixed precision training & \ttt{fp16} \\
    Random seeds & $\{42, 43, 44\}$ \\
    Learning rates & $\{1e^{-5}, 3e^{-5}, 5e^{-5}\}$ \\
    Batch sizes & $\{32, 64, 256\}$ \\
    Training steps &$\{8000,15000,30000\}$\\
    \bottomrule
    \hline
    \end{tabular}
    }
    \caption{
    {Hyperparameters in \ours{} fine-tuning.} 
    }
    \label{tab:ft_hp}
    \end{table}

\begin{table}[h!]
    \centering
    \scalebox{0.92}{
    \begin{tabular}{ll}
    \hline
    \toprule
    \tf{Task} & \tf{Label Text} \\
    \midrule
    MNLI & \ttt{yes,maybe,no}\\
    QNLI, QQP, MRPC, RTE, CoLA & \ttt{yes,no}\\
    SST-2 & \ttt{good,bad}\\
    STS-B & \ttt{0.0,0.2,0.4,...}\\
    \bottomrule
    \hline
    \end{tabular}
    }
    \caption{
    {Label text for the sequence-to-sequence setting of GLUE fine-tuning.} 
    }
    \label{tab:seq2seq_label}
    \end{table}

\section{Autoregressive Screenshot LMs}
\label{app:auto}

For both the train-from-scratch and the fine-tuning-from-Sheared-LLaMA settings,
we set each training instance to consist of 512 image patches (rendered from a sequence of 256 text tokens) 
and its subsequent 256 text tokens.
We also train text baselines with the same configuration, except that its inputs are sequences of 512 text tokens.

\Cref{tab:auto_arch} shows the configurations of our 380M models and our 1.3B models.
Both the screenshot versions and the text-only versions follow these configurations.
Note that the 1.3B configuration follows~\citet{xia2023sheared},
as we fine-tune the models from Sheared-LLaMA-1.3B.

\begin{table}[h!]
    \centering
    \scalebox{0.92}{
    \begin{tabular}{lll}
    \hline
    \toprule
    \multicolumn{1}{l}{\multirow{-1}{*}{\textbf{Component}}} & \textbf{380M} & \textbf{1.3B} \\
    
    \midrule
    Image patch size & 16 $\times$ 16 & $16 \times 16$ \\
    Hidden size & 1024 & 2048 \\
    Intermediate size & 2816 & 5504 \\
    \#Attention heads & 16 & 16 \\
    \#Layers & 24 & 24 \\    
    \bottomrule
    \hline
    \end{tabular}
    }
    \caption{{Architecture configurations of our autoregressive models.} }
    \label{tab:auto_arch}
    \end{table}

We follow most of the settings from the encoder-decoder experiments:
we use FlashAttention to speedup training;
we render the text with Google Noto Sans font with font size 10px; %
we replace all the newline symbols with \ttt{////}; 
we set an attention mask to avoid attending to the white patches at the end; 
we normalize the pixel inputs and standardize the target pixels in each patch before calculating the MSE loss.
There are also several differences: 
we do not use the prefix text or the end-of-sequence black patch,
as we find autoregressive training is more stable
and does not require these design choices.

\Cref{tab:auto_hp} shows the hyperparameters used for training our 380M and 1.3B models.
Note that the data for the two settings are different:
we use Wikipedia and BookCorpus for the 380M train-from-scratch setting,
and RedPajama~\citep{together2023redpajama} for the fine-tuning from Sheared-LLaMA setting.
In both cases,
each training instance contains 512 text tokens;
for a screenshot autoregressive model, the first 256 tokens are rendered as a $16\times 8192$ image,
which leads to $512$ patch tokens with a patch size of $16\times 16$.

\begin{table}[h!]
    \centering
    \scalebox{0.92}{
    \begin{tabular}{ll}
    \hline
    \toprule
    \multicolumn{1}{l}{\multirow{-1}{*}{\textbf{Parameter}}} & \textbf{Value (380M / 1.3B)} \\
    
    \midrule
    Learning rate & $1.5e^{-4}$ \\
    Minimum learning rate & $0$\\
    Warmup & $50$K / $2$K \\
    Learning rate scheduler & Cosine decay  \\
     Batch size & $256$ \\
    Optimizer & AdamW  \\
    Mixed precision training & \texttt{fp16} \\
    Number of steps & $16$ epochs (roughly $500$K steps) / $50$K steps\\    
    \bottomrule
    \hline
    \end{tabular}
    }
    \caption{
    {Hyperparameters in autoregressive screenshot LM training.} 
    }
    \label{tab:auto_hp}
    \end{table}

\section{More Results}
\label{app:moreresults}

\begin{figure}[h!]
    \center
    \includegraphics[width=0.98\textwidth]{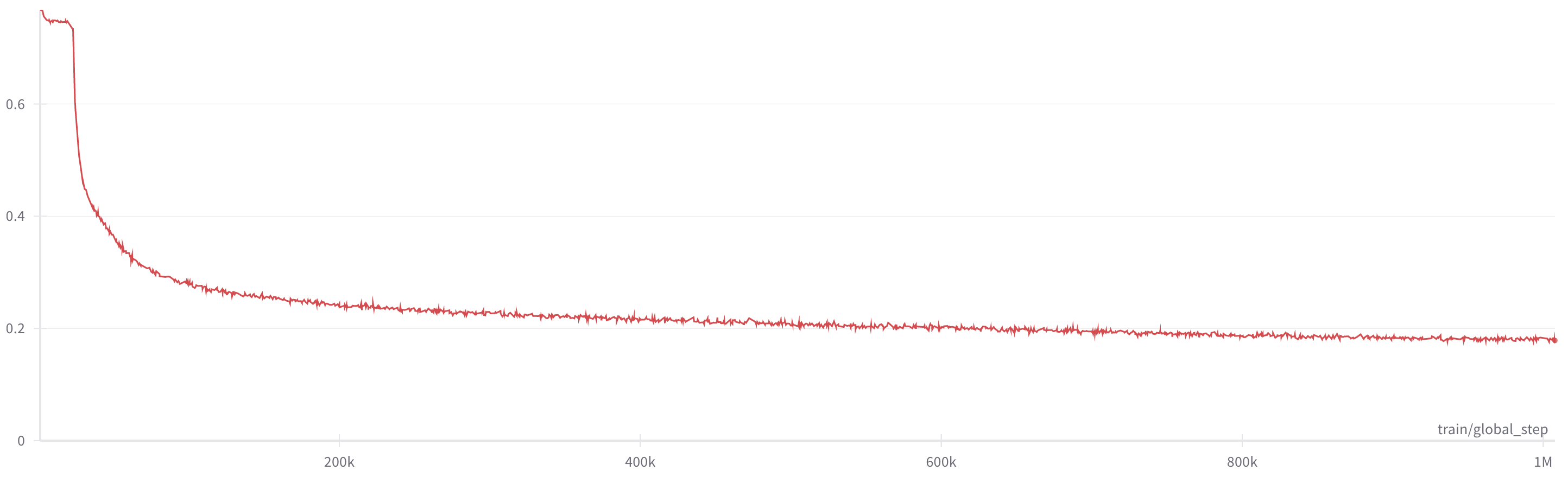}
    \caption{The patch prediction loss curve of our \ours{} model.}
    \label{fig:maeloss}
\end{figure}

\paragraph{Training loss curve.}
\Cref{fig:maeloss} shows the image prediction loss curve of our main \ours{} model.
Distinct from text-only LMs whose loss usually drops quickly at the beginning of the training,
screenshot LMs first go through a ``plateau'' phase (roughly 20K steps) and then the loss starts to decrease.

\paragraph{Main experiments.}
\Cref{tab:main_glue_rep_error} 
shows our main GLUE results with standard deviation, averaged over three seeds.
\Cref{tab:main_glue_test} 
shows the GLUE test results
of our models. We select the best performing model on the validation set (with seed 42) 
and submit the test prediction results to the GLUE leaderboard.

We also add two models here for 
comparison: 
(1) A text-only LM baseline named \textbf{Our text LM}, 
for a more fair comparison in terms of both the amount of training data and the objective. This is a text-only encoder-decoder model with the same text objective as our screenshot LMs---we randomly mask 25\% of the text tokens, replace them with \ttt{<mask>}, and use a decoder to recover the original sequence. %
(2) A reproduced version of PIXEL (\tf{PIXEL$_\text{reproduced}$}). 
We follow the exact same recipe as \citet{rust2023language} except that 
(a) we use our own rendering engine (\Cref{app:rendering}) and rendering strategies (\Cref{sec:method_bi:rendering}), and
(b) we adopt the text prefix design as described in \Cref{sec:training_stability}.

\begin{table}[h]
    \centering
    \resizebox{0.95\textwidth}{!}{
        \begin{tabular}{lrrcccccccc}
            \hline
            \toprule
            \tf{Model}& \tf{$|\theta|$} & \tf{MNLI} & \tf{QQP} & \tf{QNLI} & \tf{SST-2} & \tf{CoLA} & \tf{STS-B} & \tf{MRPC} & \tf{RTE} \\
            \midrule
            Our text LM & 115M & 85.4$_{0.2}$ / 85.4$_{0.0}$ & 88.3$_{0.1}$ & 92.2$_{0.0}$ & 92.6$_{0.3}$ & 56.3$_{0.4}$ & 90.1$_{0.1}$ & 90.8$_{0.4}$ & 62.5$_{2.1}$ \\
            Our text LM$_\text{s2s}$ & 297M & 86.0$_{0.1}$ / 86.0$_{0.3}$ & 88.5$_{0.0}$ & 92.4$_{0.1}$ & 92.7$_{0.2}$ & 58.8$_{1.5}$ & 89.3$_{0.1}$ & 91.6$_{0.3}$ & 74.2$_{0.9}$ \\
            \midrule
 {\bi{}} & 86M& 80.9$_{0.1}$ / 81.1$_{0.1}$ & 87.4$_{0.2}$ & 89.6$_{0.2}$ & 92.0$_{0.3}$ & 45.7$_{1.4}$ & 87.2$_{0.2}$  &  89.7$_{0.3}$  & 68.7$_{0.2}$\\
 {\bi{}}$_\text{s2s}$ & 268M & 82.2$_{0.1}$ / 82.6$_{0.1}$ & 87.7$_{0.1}$ & 90.4$_{0.0}$ & 92.5$_{0.2}$ &  48.8$_{0.7}$ & 83.8$_{0.2}$  & 90.6$_{0.2}$ & 67.7$_{0.3}$ \\

        \bottomrule
        \hline
        \end{tabular}
    }
    \caption{
        GLUE validation results with standard deviation.
    }
    \label{tab:main_glue_rep_error}
\end{table}

\begin{table}[h]
    \centering
    \resizebox{0.8\textwidth}{!}{
        \begin{tabular}{lrccccccc}
            \hline
            \toprule
            \tf{Model}&  \tf{MNLI} & \tf{QQP} & \tf{QNLI} & \tf{SST-2} & \tf{CoLA} & \tf{STS-B} & \tf{MRPC} & \tf{RTE} \\
            \midrule
 BERT & 84.6/83.4 & 71.2 & 90.5 & 93.5 & 52.1 & 85.8 & 88.9 & 66.4\\
 \midrule
 PIXEL$_\text{reproduced}$ & 73.5/73.7 & 69.7 & 85.0 & 88.7 & 9.9 & 79.4 & 85.1 & 56.8\\
 \bi{} & 79.9/79.4 & 69.8 & 88.4 & 90.0 & 37.9 & 79.4 & 85.7 & 62.0\\
 \bi{}$_\text{s2s}$ & 81.8/80.9 & 70.5 & 89.3 & 91.6 & 43.1 & 87.9 & 87.5 & 61.7\\

        \bottomrule
        \hline
        \end{tabular}
    }
    \caption{
        GLUE test results.
}        
    \label{tab:main_glue_test}
\end{table}

\vspace{20pt}

\paragraph{Embedding layernorm ablation.}
\Cref{tab:layer_norm_table}
shows the effect of using embedding layernorm 
on models with different masking rates.
Note that the training stability results may vary depending on the hardware, software, and random seeds used. 
As the table has demonstrated,
higher masking rates or larger patch sizes lead to an increased chance
of training instability. 
Using embedding layernorm can prevent loss spikes in most cases.

\begin{table}[t!]
    \centering
    \resizebox{0.65\textwidth}{!}{
        \begin{tabular}{ccccc}
            \hline
            \toprule
            \multirow{2}{*}{\tf{Patch Size}} & \multicolumn{2}{c}{\tf{Mask Rate}} & \multirow{2}{*}{\tf{w/o Embedding LN}} & \multirow{2}{*}{\tf{w/ Embedding LN}} \\
            \cmidrule{2-3}
            \phantom{} & Patch & Text & \phantom{} & \phantom{} \\
            \midrule
            \multirow{4}{*}{$16\times16$} & 10\% & 25\% & \cmark & \cmark \\
            \phantom{} & 10\% & 40\% & \cmark & \cmark \\
            \phantom{} & 25\% & 25\% & \cmark & \cmark \\
            \phantom{} & 25\% & 40\% & \xmark & \cmark \\
            \midrule
            $16\times32$ & 10\% & 25\% & \xmark & \cmark \\
            \midrule
            $16\times64$ & 10\% & 25\% & \xmark & \cmark \\
            \bottomrule
            \hline
        \end{tabular}
    }
    \caption{
        Effects of using embedding layernorm.
        \cmark{} indicates that the training can be completed smoothly while
        \xmark{} indicates that the training collapsed due to loss spikes.
        Note that this result may vary depending on the hardware, software versions, and random seeds.
    }
    \label{tab:layer_norm_table}
\end{table}

\paragraph{Span masking.}
\Cref{tab:ablation_spanmasking} shows the comparison between using span masking vs. not using span masking on image patches. 
We observe that at this masking rate, using span masking for image patches leads to a significantly better result.

\begin{table}[h!]
    \centering
    \resizebox{0.48\linewidth}{!}{
        \begin{tabular}{cccccc}
            \hline
            \toprule
            {\textbf{Span Masking}} & \tf{MNLI} &{\tf{SST-2}} & \tf{MRPC} &{\tf{RTE}}\\
            \midrule 
            \colorcello \cmark & \colorcello \tf{80.9} & \colorcello \tf{92.0} & \colorcello \tf{89.7} & \colorcello \tf{68.7} \\
            \xmark & 80.5 & 90.9 & 89.3 & 64.9 \\
        \bottomrule
        \hline
        \end{tabular}
    }
    \caption{
        Ablation on span masking. Both models use a 10\% patch masking rate and a 25\% text masking rate.
    }
    \label{tab:ablation_spanmasking}
\end{table}

\clearpage

\section{Examples}
\label{app:example}

\begin{figure*}[h]
    \centering
    \includegraphics[width=0.98\linewidth]{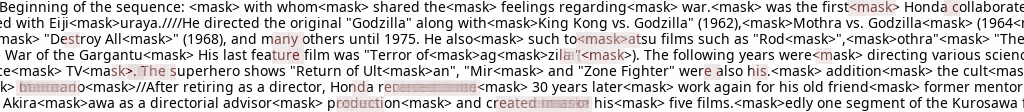}
    \caption{The prediction of our main model at 100K training step.}
    \label{fig:exp_1025_100k}
\end{figure*}

\begin{figure*}[h]
    \centering
    \includegraphics[width=0.98\linewidth]{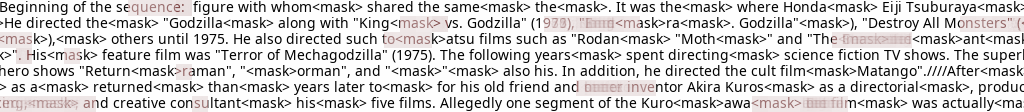}
    \caption{The prediction of our main model at 500K training step.}
    \label{fig:exp_1025_500k}
\end{figure*}

\begin{figure*}[h]
    \centering
    \includegraphics[width=0.98\linewidth]{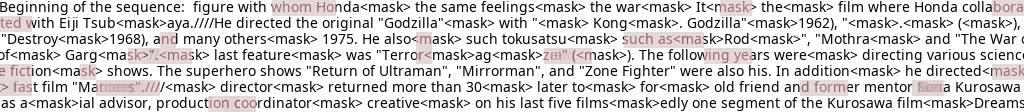}
    \caption{The prediction of our main model at 1M training step.}
    \label{fig:exp_1025_1m}
\end{figure*}

\Cref{fig:exp_1025_100k},
\Cref{fig:exp_1025_500k}, and
\Cref{fig:exp_1025_1m}
show how the prediction of our main model (10\%span/25\%) evolves during pre-training.
The model becomes more and more capable of predicting longer masked spans.


\begin{figure*}[h!]
    \centering
    \includegraphics[width=0.98\linewidth]{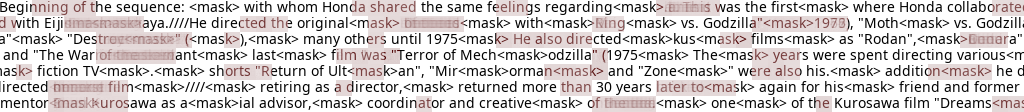}
    \caption{The prediction of the 25\%span/25\% (patch/text) masking model with $16\times16$ patch size.}
    \label{fig:exp_25span_25}
\end{figure*}

\begin{figure*}[h!]
    \centering
    \includegraphics[width=0.98\linewidth]{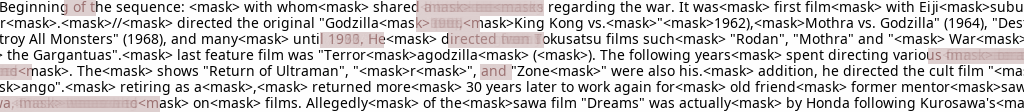}
    \caption{The prediction of the 10\%span/25\% (patch/text) masking model with $16\times32$ patch size.}
    \label{fig:exp_10251_16x32}
\end{figure*}

\Cref{fig:exp_25span_25} shows the prediction of the 25\%span/25\% (patch/text) masking model with $16\times16$ patch size. With more masking, accurately predicting the masked patches becomes increasingly difficult.
\Cref{fig:exp_10251_16x32} shows the prediction of the 10\%span/25\% (patch/text) masking model with $16\times32$ patch size. The larger patch size essentially leads to more ``span'' masking,
making the pre-training task more challenging.

\end{document}